\documentclass[10pt,twocolumn,letterpaper]{article}
\usepackage[pagenumbers]{cvpr}

\usepackage{cite}
\usepackage{amsmath,amssymb,amsfonts}
\usepackage{algorithmic}
\usepackage{graphicx,color}
\usepackage{textcomp}
\usepackage{xcolor}
\usepackage{hyperref}
\hypersetup{hidelinks=true}
\usepackage{algorithm,algorithmic}
\def\BibTeX{{\rm B\kern-.05em{\sc i\kern-.025em b}\kern-.08em
    T\kern-.1667em\lower.7ex\hbox{E}\kern-.125emX}}
\AtBeginDocument{\definecolor{tmlcncolor}{cmyk}{0.93,0.59,0.15,0.02}\definecolor{NavyBlue}{RGB}{0,86,125}}

\usepackage{booktabs}

\usepackage{soul} %
\usepackage{array}
\usepackage[capitalize]{cleveref}

\usepackage[capitalize]{cleveref}
\crefname{section}{Sec.}{Secs.}
\Crefname{section}{Section}{Sections}
\Crefname{table}{Table}{Tables}
\crefname{table}{Tab.}{Tabs.}

\usepackage[toc,page,titletoc]{appendix}

\usepackage{array}

\begin{document}

\title{Tell Me What You See: Text-Guided Real-World Image Denoising}

\author{Erez Yosef\\
Tel Aviv University, Israel\\
{\tt\small erez.yo@gmail.com}
\and
Raja Giryes\\
Tel Aviv University, Israel\\
{\tt\small raja@tauex.tau.ac.il}
}
\maketitle

\begin{abstract}
Image reconstruction from noisy sensor measurements is challenging and many methods have been proposed for it. Yet, most approaches focus on learning robust natural image priors while modeling the scene’s noise statistics. In extremely low-light conditions, these methods often remain insufficient. Additional information is needed, such as multiple captures or, as suggested here, scene description. As an alternative, we propose using a text-based description of the scene as an additional prior, something the photographer can easily provide. Inspired by the remarkable success of text-guided diffusion models in image generation, we show that adding image caption information significantly improves image denoising and reconstruction for both synthetic and real-world images. All code and data will be made publicly available upon publication.

\end{abstract}

\maketitle

\section{Introduction}
\label{sec:intro}

Image denoising and reconstruction are fundamental problems in imaging, with numerous approaches proposed to address them \cite{Ongie2020Deep}.
In scenes with poor lighting conditions or other limiting factors such as short exposure intervals (e.g. for dynamic scenes), the signal-to-noise ratio (SNR) is significantly reduced, resulting in noisy sensor measurements. Image denoising under these conditions remains challenging, and despite prior research, image acquisition in such constrained scenarios is still difficult.

Since the true noise statistics are unknown and camera-specific, various methods have been suggested to better approximate noise characteristics. Early work relied on Gaussian noise models, which were later improved with the Poisson-Gaussian noise model \cite{foi2008practical}. To reconstruct the true signal from noisy measurements numerous image priors were proposed \cite{Ongie2020Deep,gilton2021model}.
Yet, under severe noise conditions, the reconstruction task becomes highly ill-posed and the available natural image priors are not specific and informative enough. Consequently, more information about the scene is essential to enhance reconstruction performance.

\begin{figure}[t]
    \setlength\tabcolsep{1pt} %
    \centering
    \begin{tabular}{ccc}
       
        \footnotesize Input image & \footnotesize Diffusion without text & \footnotesize \shortstack{Diffusion with text} \\
        \midrule
        \includegraphics[width=0.310\linewidth]{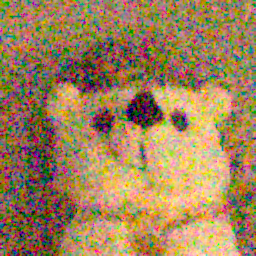} & \includegraphics[width=0.31\linewidth]{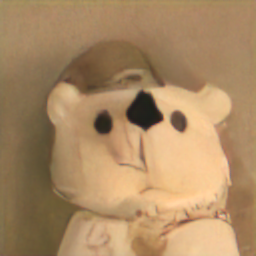}  & \includegraphics[width=0.31\linewidth]{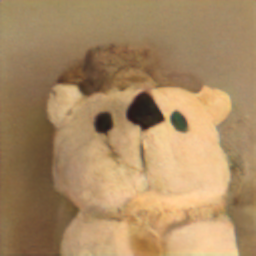} \\
        \footnotesize \shortstack{Noisy raw capture\vspace{3.2 mm}} & \footnotesize \shortstack{No text  \vspace{3.2 mm}} & \footnotesize \textit{\shortstack{A fluffy furry hedgehog\\doll in brown colors}} \\
                \includegraphics[width=0.310\linewidth]{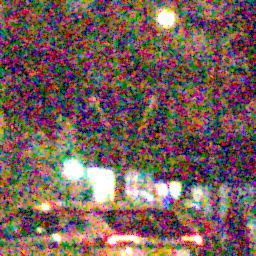} & \includegraphics[width=0.31\linewidth]{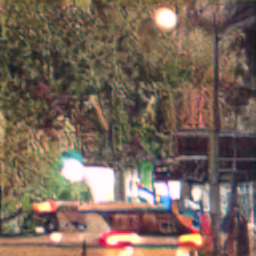} & \includegraphics[width=0.31\linewidth]{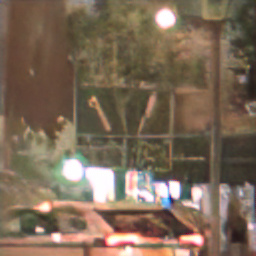} \\
        \footnotesize \shortstack{Noisy raw capture \vspace{3.2 mm} } & \footnotesize \shortstack{No text \vspace{3.2 mm}} & \footnotesize \textit{\shortstack{A green road sign with two \\ white arrows on the street}} \\   
    \end{tabular}
    \vspace{-0.1in}
    \caption{Raw noisy images captured with a smartphone camera (left) were reconstructed using diffusion models, both without (center) and with (right) a text caption. The contribution of the text description to the reconstruction and perceptual quality is significant.}
    \label{fig:teaser}
\end{figure}

\definecolor{cy}{rgb}{1, 0.95, 0.8}%
\definecolor{cb}{rgb}{0.92, 0.94, 1} 
\newcommand{\hlc}[2][yellow]{{%
    \colorlet{foo}{#1}%
    \sethlcolor{foo}\hl{#2}}%
}

\begin{figure*}[t]
\begin{center}
   \includegraphics[width=\linewidth]{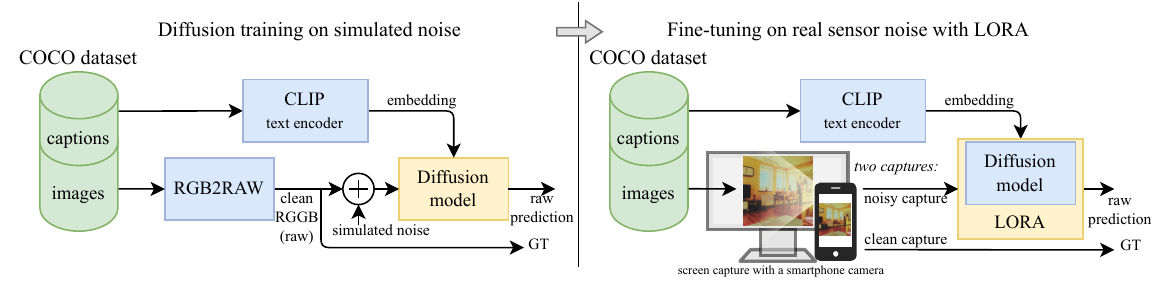}
\vspace{-0.2in}
   \caption{Proposed training framework: Initially (left), we train a diffusion model on the COCO-captions dataset \cite{chen2015microsoft} with simulated noise. Then (right), we fine-tune the model for real-world noise on screen-captured images. Each sample is captured twice with different camera settings, resulting in noisy and clean image pairs for training. In this scheme, \hlc[cb!150]{\textbf{blue}} blocks weights are pre-trained and fixed, while \hlc[cy!120]{\textbf{yellow}} blocks weights are trained in each stage.}
   \label{fig:system_diagram}
\end{center} 
\end{figure*}

Following this observation, we propose using a text caption to aid in reconstructing the captured scene. This additional information can be provided by the photographer. To integrate the caption into the reconstruction process, we employ a text-guided diffusion model, leveraging the advances in this field for image reconstruction and generation \cite{rombach2022high, nichol2021glide, dhariwal2021diffusion, kawar2022denoising, saharia2022palette, croitoru2023diffusion}. A multimodal model, CLIP \cite{radford2021learning}, is used to combine the caption and its corresponding raw image into a unified reconstruction framework.

We train a diffusion model in the raw image domain using a dataset of images paired with text captions. Initially, we trained on simulated noise. However, since the simulated noise model does not accurately simulate true sensor noise, and our goal is real-world imaging, we fine-tune the trained model on real-world noisy data to enhance its performance. For this fine-tuning, a low-rank set of model weights is optimized with LoRA \cite{hu2021lora} using a small dataset containing clean and noisy image pairs with corresponding text captions, captured efficiently using a target camera and a computer screen (\Cref{sec:method}). \Cref{fig:system_diagram} illustrates our capturing and training scheme for fine-tuning with real-world data.

Our approach improves reconstruction results compared to other non-text-guided methods, including a similar diffusion model without text guidance (\Cref{fig:teaser}). We evaluate and demonstrate the superiority of our approach on both synthetic and real-world captured images. For fine-tuning and evaluation, we tested two target cameras separately: Samsung S21 (world-facing camera) and Allied Vision Manta G-146C.

The contributions of our work are as follows:
(i) A novel method for raw image capturing and reconstruction based on a text-guided raw diffusion model. This approach leverages caption information through recent multimodal methods and advanced processing capabilities. 
(ii) State-of-the-art performance both quantitatively and qualitatively (\Cref{sec:results}) compared to previous methods, including for outdoor real-world imaging scenarios.
(iii) A novel real-noise captioned dataset, containing pairs of clean and real sensor noisy raw images with corresponding text captions. This dataset is specifically designed to advance research in this domain.
(iv) Datasets, Codes, and models will be made publicly available upon acceptance.

\section{Background and Related Work}
\label{sec:relatedwork}

Classical image denoising methods, such as thresholding \cite{donoho1995noising} and total variation \cite{rudin1992nonlinear}, used hand-crafted parametric algorithms to recover the clean image. These methods heavily relied on assumptions about the data and noise statistics.

Due to the limitations of these algorithms, non-parametric and self-similarity algorithms were proposed, such as BM3D \cite{dabov2007image}, non-local means \cite{buades2005non}, dictionary learning \cite{aharon2006k} and Field-of-Experts \cite{Roth2005Fields}. 
Nowadays, most denoising algorithms are data-driven and leverage deep neural networks. Following the pioneering work that trained Multi-Layer Perceptron (MLP) on large synthetic noise images \cite{burger2012image}, several deep learning methods have been introduced to further enhance performance \cite{Remez2018Class,izadi2023image, guo2019toward, gharbi2016deep, Schwartz2019DeepISP, brooks2019unprocessing, anwar2019real, zhang2017beyond, zhang2018ffdnet, zhao2022hybrid, huang2022towards, jin2023dnf, zamir2020cycleisp,zamir2021multi,tu2022maxim, zamir2022restormer, chen2022simple}.

Due to the statistical differences between simulated noise and real sensor noise, having a real camera-captured dataset is essential for improving real-world denoising performance. Several datasets of clean and noisy image pairs were captured and aligned, such as the SIDD \cite{abdelhamed2018high} dataset for smartphone cameras and DND \cite{plotz2017benchmarking} for consumer cameras, both providing raw and sRGB data for benchmarking.
Capturing clean and noisy image pairs is not trivial as careful alignment is required, and both the camera and the scene must remain static.
To overcome the challenges of real data acquisition, self-supervised methods have been proposed that use only noisy samples without requiring ground truth \cite{krull2019noise2void, batson2019noise2self, huang2021neighbor2neighbor, ulyanov2018deep, Quan2020Self2Self, zhang2022idr, zhang2023self}.

\noindent \textbf{Denoising Diffusion Models} \cite{ho2020denoising,nichol2021improved,croitoru2023diffusion}, are generative models that have gained large popularity in recent years due to their strong capabilities in image generation \cite{rombach2022high, nichol2021glide, dhariwal2021diffusion}, segmentation \cite{amit2021segdiff, baranchuk2021label} and reconstruction \cite{kawar2022denoising, saharia2022palette, lugmayr2022repaint, batzolis2021conditional, saharia2022image, abu2022adir, zhu2023denoising, fei2023generative,croitoru2023diffusion}.
Diffusion models utilize a parameterized Markov chain to generate samples of a data distribution after several steps. In the forward direction, the Markov chain gradually adds noise to the data until it is mapped to a simple distribution (an isotropic Gaussian). To sample an image, we reverse the Markov chain, starting with a pure noise image from the isotropic Gaussian distribution and progressively denoise the image using a trained deep network.

In the context of low-level image restoration, diffusion models have been applied to various tasks, including image restoration for linear inverse problems \cite{kawar2022denoising, chung2023prompt, delbracio2023inversion}, spatially-variant noise removal \cite{pearl2023svnr}, and low-light image enhancement and denoising using physical approaches \cite{yi2023diff, torem2023complex}.
Additionally, diffusion models have been employed for low-light text recognition \cite{nguyen2023diffusion}.
Some concurrent works have presented image processing by text instructions \cite{qi2023tip, yan2023textual}, a latent diffusion inverse solver incorporating regularization by text \cite{kim2023regularization}, text-guided super-resolution \cite{chen2023image}, and text manipulations for image reconstruction \cite{lin2024improving, yu2024scaling}.
While their use of text is similar to our approach, they primarily focus on RGB processing methods and known degradation models, whereas our method targets low-level denoising of raw images with the necessary training modifications for raw data and an unknown sensor noise distribution. \Cref{sec:method} shows that raw denoising is superior to RGB denoising.

Our approach uses denoising diffusion probabilistic models (DDPM). We now describe the DDPM parts relevant to our process. For more details see \cite{nichol2021improved, ho2020denoising, dhariwal2021diffusion}.
The clean input data $x_0 \sim q(x_0)$ is drawn from the distribution $q$, and the latent steps of the process are $x_1,x_2,\ldots,x_T$ (for T timesteps). At each time step, the forward process is
\begin{equation}\label{eq:forward_diffusion}
q(x_{t}|x_{t-1}):=N(x_{t};\sqrt{1-\beta _{t}}x_{t-1},\beta _{t}\mathbf{I}),
\end{equation}
i.e., a noisier sample is generated given the previous step sample, where $\beta_{1},...,\beta_{T}$ is a fixed variance schedule of the process designed such that approximately $x_T \sim N(0,I)$.

A key property of the forward process is that sampling $x_t$ at any timestamp $t$ given $x_0$ can be expressed as
\begin{equation}\label{eq:property_derivation}
q(x_{t}|x_{0}):=N(x_{t};\sqrt{\overline{\alpha}_{t}}x_{0},(1-\overline{\alpha}_{t})\mathbf{I}),
\end{equation}
where $\alpha _{t}:=1-\beta _{t}$, $\overline{\alpha}_{t}:=\prod_{s=1}^t \alpha _{s}$.
Thus, $x_{t}$ can be expressed as a linear combination of $x_{0}$ and noise $\epsilon \sim N(0,\mathbf{I})$:
\begin{equation}\label{eq:property}
x_{t}=\sqrt[]{\overline{\alpha} _{t}}x_{0}+\sqrt[]{1-\overline{\alpha} _{t}}\epsilon.
\end{equation}
In the reverse process, a signal is iteratively recovered from the noise. The previous timestamp sample $x_{t-1}$ is achieved using a trained network. The sample at $t-1$ can be described as a Gaussian with learned mean and fixed variance \cite{ho2020denoising}:
\begin{equation}\label{eq:reverse_diffusion}
p_{\theta}(x_{t-1}|x_{t})=N(x_{t-1};\mu_{\theta}(x_{t},t),{\sigma}_t^2\mathbf{I}).
\end{equation}
Diffusion model can be conditioned by additional information $y$, so the data conditional distribution is $x_0 \sim q(x_0|y)$, and the reverse step model takes $y$ as an additional input $\mu_{\theta}(x_{t}, y,t)$ to get a conditional prediction. %

\noindent \textbf{Evaluation metrics.} Note that generative models, such as diffusion models, are mainly designed to learn a representative posterior distribution that maximizes perceptual quality, rather than a deterministic solution that reduces the L2-norm and induces high PSNR \cite{theis2015note}.
This difference is discussed and analyzed in the context of the `perception-distortion tradeoff' \cite{blau2018perception} and has been the subject of research aiming to improve perceptual quality assessment \cite{gu2020pipal, gu2022ntire}. 
As generative models prioritize perceptual quality, they perform worse on traditional distortion metrics such as PSNR and SSIM \cite{wang2004image} which do not capture perceptual quality \cite{wang2004image, wang2009mean}, and higher values of these metrics may not necessarily correlate with improved perceptual quality \cite{ledig2017photo, zhang2018perceptual}. Consequently, we employ the LPIPS \cite{zhang2018perceptual} and DISTS \cite{ding2020iqa} metrics to provide a more reliable perceptual evaluation.

\noindent \textbf{Low-Rank Adaptation (LoRA)} \cite{hu2021lora} is a method for fine-tuning a network trained for a general task to adapt it to a new task. Originally introduced for large language models, it has also been applied to diffusion models \cite{seo2023let, lim2023image}. In LoRA, a low-rank weight matrix is added to the pre-trained weights of the network, and only this small set of parameters is fine-tuned, while the original large network remains fixed.

\noindent \textbf{Text Guidance.}
Models such as CLIP map images and text jointly into a shared space \cite{radford2021learning} using an image encoder and a text encoder that are trained such that the dot product between the representation vectors of a corresponding image-text pair is maximized. As a result, matching text and image pairs get a high score, while non-corresponding pairs get a low dot product score. This enables the combination of the two domains, e.g,
using text guidance for image generation \cite{nichol2021glide, saharia2022photorealistic, ramesh2022hierarchical}, editing \cite{kawar2023imagic,bai2023textir,li2020manigan,brooks2023instructpix2pix,Valevski2023UniTune} and image segmentation \cite{luddecke2022image}. We are the first to use text guidance for real-world denoising in the raw image domain, which requires specialized datasets containing raw images with real sensor noise and corresponding captions that do not exist to date. To address this gap, we provide the necessary real-world data for this task, along with a dedicated training and fine-tuning scheme.

\section{Method}
\label{sec:method}

We designed and trained a text-guided diffusion model to denoise noisy raw images, given a text description of the captured scene. We apply the diffusion model in the sensor raw domain rather than in the RGB space since transforming to RGB requires spatial and non-linear operations, which makes the noise more complex and challenging to handle.

We next describe our strategy for noise modeling and data processing and then our text-guided denoising scheme. %

\noindent \textbf{Noise Modeling.} %
The digital imaging process is affected by intrinsic noise that influences the measurements of the image sensor. A low SNR is caused by the low signal intensity, namely, the light collected by the sensor photodiodes.
Formally, the captured raw sensor image $y$ can be represented as random variables with a distribution conditioned on the true scene image $z$, denoted as $p_{cam}(y|z)$.
Camera noise primarily stems from two components: photon arrival statistics and readout circuitry noise. Photon noise (shot noise) follows a Poisson distribution with a mean equal to the true light intensity. The readout noise distribution can be approximated by Gaussian noise with zero mean and fixed variance \cite{foi2008practical}.

Data for supervised training of deep learning models for raw image denoising tasks is difficult to obtain, as pairs of clean and noisy images are required. This requires capturing each pair sequentially with different settings to control the noise levels. Additionally, while noise can be reduced using appropriate imaging setups, it cannot be entirely eliminated, as noise is an intrinsic part of the imaging process, and acquiring a true clean image is not practical.

To obtain a large dataset with noisy and clean images, we simulate camera noise using a noise model from prior works on raw image denoising \cite{brooks2019unprocessing, zamir2020cycleisp}, where the overall noise can be approximated as a heteroscedastic Gaussian with variance dependent on the true image $z$:
\begin{equation}\label{eq:noise_gauss_model}
y \sim p_{cam}(y|z) \approx N(y;\mu=z,\sigma^2=\lambda_{read}+\lambda_{shot}z),
\end{equation}
where the variance parameters $\lambda_{read}$ and $\lambda_{shot}$ are determined based on the sensor's analog and digital gains.
Following the real-world sensor noise statistics from \cite{brooks2019unprocessing}, we sample the noise level parameters $\lambda_{read}$ and $\lambda_{shot}$ of the read and shot components from the distributions
\begin{align}
\label{eq:noise_lambda_params_sample}
\log(\lambda_{shot}) &\sim U(a=\log(0.1), b=\log(0.31)),\\
\begin{split}
\log(\lambda_{read}) \;| \;\log(&\lambda_{shot}) \sim\\ N(\mu&=1.5\cdot \log(\lambda_{shot})+0.05,\sigma^2=0.5).
\end{split}
\end{align}
The range of these noise parameters is chosen to align with the noise statistics of real cameras, as in \cite{brooks2019unprocessing}, with the range for $\lambda$ selected to be broad for robustness against noise and varying imaging conditions.
Based on this sampling scheme, we generate simulated noise for the diffusion model training.

\noindent \textbf{Data Processing.} %
Our text-based image enhancement approach requires a large dataset of raw images and corresponding text descriptions for training. We used the COCO-captions dataset \cite{chen2015microsoft} which contains approximately 120k RGB images paired with text captions. To obtain raw sensor data from the RGB images, we leveraged the RGB2RAW network presented in CycleISP \cite{zamir2020cycleisp} to convert the RGB images into sensor raw images in Bayer pattern format.
The dataset's text captions were processed using the CLIP (ViT-L/14) text encoder \cite{radford2021learning} resulting in representation vectors of length 768.
After this preprocessing, we obtained a dataset consisting of clean raw sensor images and the corresponding CLIP representations of the captions required for training.

\noindent \textbf{Diffusion Model and Training.} %
Our diffusion model was trained according to the conventional training scheme for diffusion models.
We set $T=1000$ diffusion steps with a cosine noise scheduler.
To condition the diffusion process on the noisy image, the raw noisy image $y$ is concatenated to the diffusion sample $x_t$ in each diffusion step.
We used a U-Net architecture with 8 input channels for $x_t$ and $y$ images, while each was rearranged to 4-channel RGGB format. The network output is the conditional estimation of the sample mean in 4 channels of RGGB raw image. The U-Net consists of 5 scales with 3 residual blocks in each scale. The number of output channels of the blocks in scales 1, 2, 3, 4 and 5 is 128 multiplied by $[1,1,2,3,4]$ respectively. All samples have the same spatial dimension of 256x256.

The U-Net architecture is controlled by the timestep value $t$ through positional encoding followed by two fully connected (FC) layers separated by an activation function. The network is conditioned on the text input by applying two FC layers to the CLIP text embedding vectors. The resulting vectors are summed and added to the features at each convolution block within the network. 
The model is trained to estimate the denoised sample $x_0$ using the L1 loss. 

For a baseline comparison, we trained a diffusion model without text guidance, replacing the image-specific CLIP embedding vector with a global embedding vector. The diffusion model was trained for 1.2M steps using a batch size of 16 and AdamW \cite{loshchilov2017decoupled} with a learning rate of $10^{-4}$.

\noindent \textbf{Real Camera Noise Fine-Tuning.} %
Although simulated noise was designed to closely replicate real sensor noise, discrepancies remain. Also, sensor noise characteristics are specific to each camera. Thus, fine-tuning the model for the target camera is essential for achieving improved performance.

\begin{figure}[t]
    \begin{center}
   \includegraphics[width=0.5\linewidth]{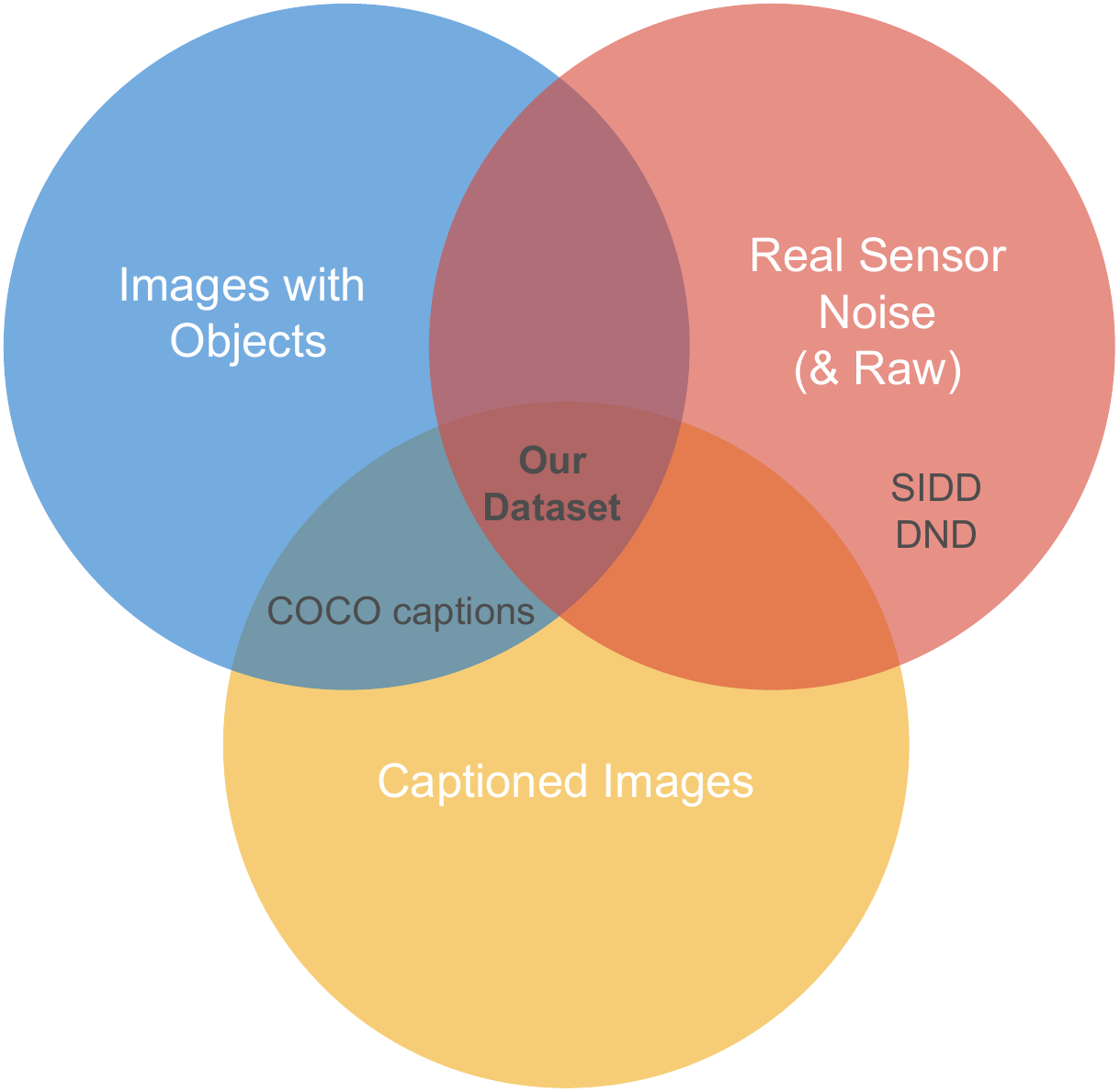}
    \caption{Our dataset consists of raw noisy images containing objects and their corresponding captions. Existing raw noisy image datasets (e.g., SIDD \cite{abdelhamed2018high} and DND \cite{plotz2017benchmarking}) include patches from large images that lack captionable content. In contrast, caption datasets (e.g., COCO-captions \cite{chen2015microsoft}) contain RGB images with captions but do not include raw noisy sensor images. Therefore, our proposed dataset is novel and contributes to text-guided real-world image denoising tasks.}
    \label{fig:venn_diagram}
    \end{center}
\end{figure}

To bridge the gap between the simulated training dataset and real-world noise statistics, which can degrade performance, we fine-tuned our trained model using real noisy sensor measurements consisting of pairs of noisy and clean images. Existing denoising datasets primarily consist of image crops, where each crop contains only a specific pattern or structure, lacking captionable content or identifiable objects. Due to the lack of a relevant dataset for our task (see \Cref{fig:venn_diagram}), which contains both raw noisy images with text captions of actual objects, we created a new dataset. We captured images using two cameras: the world-facing camera of a Samsung S21 smartphone and the Allied Vision Manta G-146C camera. The images were recorded in raw format (Bayer pattern) with 10-bit  and 12-bit depth , respectively.

The cameras were positioned in front of a LED screen displaying samples from the COCO-captions dataset \cite{chen2015microsoft}. This setup ensured that the original captions of the dataset were still relevant to the captured images without requiring re-captioning. Using the Samsung S21 camera, we captured 500 images for training and 30 for testing, while the Allied camera was used to capture 1000 images for training and 100 for testing. Each sample was captured twice: once with settings for high noise and once with settings for minimal noise, resulting in realistic noisy-clean image pairs.

To fine-tune the model on the captured data, we use LoRA \cite{hu2021lora}. It requires less data for training by optimizing a low-rank set of parameters while preserving information learned during the pretraining on large data of simulated noise. 
Low-rank weights were added to the FC layers in the attention modules, and to the convolutions of the residual blocks.

Following \cite{hu2021lora}, for the FC layers, a low-rank matrix $\Delta W$ of rank $r$ was added to the pre-trained matrix weight $W_0\in R^{d\times k}$.
The low-rank matrix  $\Delta W$ is decomposed as $\Delta W = AB$, where $A\in \mathbb{R}^{d\times r}$ and  $B\in \mathbb{R}^{r\times k}$ are trainable matrices. Given an input $x$ and output $y$ the layer’s operation is 
\begin{equation}\label{eq:lora_weight}
y= W_0x + \Delta Wx = W_0x + AB x,
\end{equation}

For 2D convolutions, the low-rank operation is applied along the channel dimension. Specifically, for a convolution layer with $d$ input and $k$ output channels (features), an additional convolution layer `A' (with the same kernel size as the pre-trained convolution) mapped the $d$ input channels to $r$ channels, which were then remapped to $k$ output channels via a 1×1 convolution layer `B'. The low-rank bypass is summed with the pre-trained layer’s output, analogous to the FC case. A diagram illustrating this process is provided in the sup. mat. We used $r=4$ in our models, initializing the low-rank parameters to zero in the fine-tuning process.

\section{Results}
\label{sec:results}
\newcommand{\cycle}{figs/coco_sim/cycleisp01/fix}
\newcommand{\ntov}{figs/coco_sim/n2v}
\newcommand{\noisy}{figs/coco_sim/noisy}
\newcommand{\concat}{figs/coco_sim/concat}
\newcommand{\cond}{figs/coco_sim/cond}
\newcommand{\gt}{figs/coco_sim/gt}
\newcommand{\NAFnet}{figs/coco_sim/NAFnet}
\newcommand{\restormer}{figs/coco_sim/restormer}

\newcommand{\imgA}{0} 
\newcommand{\imgB}{2} 
\newcommand{\imgC}{4} 

\newcommand{\imnum}[1]{00000#1.pt.png} 
\newcommand{\dimnum}[1]{sample00#1.png}
\newcommand{\gtimnum}[1]{gt00#1.png}

\newcommand{\imnumA}{000003.pt.png} 
\newcommand{\dimnumA}{sample003.png}

\newcommand{\imnumB}{000004.pt.png} 
\newcommand{\dimnumB}{sample004.png}
\newcommand{\imnumC}{000005.pt.png} 
\newcommand{\dimnumC}{sample005.png}
\newcommand{\nlvl}{01}

\newcommand{\pfig}[1]{\raisebox{-.5\totalheight}{\includegraphics[width=0.13\linewidth]{#1}}}

\newcommand{\ourdiffusion}{Diffusion \footnotesize(our)}
\newcommand{\ourtextcond}{+ Text \footnotesize (our)}

\begin{figure*}[t]
    \setlength\tabcolsep{2pt} %
    \centering
    \begin{tabular}{cccccm{1.9cm}c}
        Noisy 0.1 & CycleISP & Noise2Void &  Diffusion &\multicolumn{2}{|c|}{Diffusion + Text} &\hspace{0.5mm} GT \vspace{-0.1cm}\\
        & & & \footnotesize (our) &  \multicolumn{2}{|c|}{\footnotesize (our)}  & \vspace{-0.1cm}\\

        \midrule
        \pfig{\noisy\nlvl/\imnum\imgA} & \pfig{\cycle\nlvl/\imnum\imgA} & 
        \pfig{\ntov\nlvl/\dimnum\imgA} &
        \pfig{\concat\nlvl/\dimnum\imgA}& \pfig{\cond\nlvl/\dimnum\imgA} &
        {\small A room with chairs, a table, and a woman in it.} &\hspace{0.5mm}
        \pfig{\gt/\gtimnum\imgA}\\
    
        \pfig{\noisy\nlvl/\imnum\imgB} & \pfig{\cycle\nlvl/\imnum\imgB} & 
        \pfig{\ntov\nlvl/\dimnum\imgB}&

        \pfig{\concat\nlvl/\dimnum\imgB}& \pfig{\cond\nlvl/\dimnum\imgB} &
        \small Bedroom scene with a bookcase, blue comforter and window. &\hspace{0.5mm}
        \pfig{\gt/\gtimnum\imgB} \\
    
        \pfig{\noisy\nlvl/\imnum\imgC} & \pfig{\cycle\nlvl/\imnum\imgC} & 
        \pfig{\ntov\nlvl/\dimnum\imgC}&
        \pfig{\concat\nlvl/\dimnum\imgC}& \pfig{\cond\nlvl/\dimnum\imgC}&
        \small Three teddy bears, each a different color, snuggling together. &\hspace{0.5mm}
        \pfig{\gt/\gtimnum\imgC}\\
        
    \end{tabular}
    \caption{\textbf{Low simulated noise results.} Comparison of various methods for raw image denoising at a noise level of 0.1 ($\log\lambda_{shot}=0.1$ and  $\log\lambda_{read}=0.2$). Our models achieve superior performance, with text guidance significantly enhancing perceptual quality, details, and textures.}

    \label{fig:sim_result01}
\end{figure*}

\begin{figure}[!t]
    \centering
    \includegraphics[width=\linewidth]{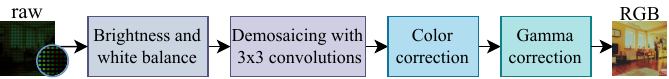}
    \vspace{-0.1in}
    \caption{The process used to convert raw images into RGB format for qualitative and quantitative evaluations.}
    \vspace{-0.15in}
    \label{fig:raw2rgb}
\end{figure}

We present both qualitative and quantitative results for simulated noise and real-world camera noise. To evaluate the obtained results relative to noise-free (ground-truth, GT) images, we measured the peak signal-to-noise ratio (PSNR) metric (on both raw and RGB images) as well as structural and perceptual metrics, including SSIM \cite{wang2004image}, LPIPS \cite{zhang2018perceptual} and DISTS \cite{ding2020iqa}, all performed on the RGB version of the images. For converting raw images to RGB format for presentation and evaluation, we utilized a deterministic and straightforward process as presented in \cite{brooks2019unprocessing} and illustrated in \Cref{fig:raw2rgb}. We trained and tested two diffusion models, the first with text guidance, and the second without text guidance. Since this is the only difference between the models, this comparison exhibits the significant contribution of text to reconstruction under high noise conditions. 

In addition, we compare our methods to previous state-of-the-art works including raw images denoising: CycleISP \cite{zamir2020cycleisp}, deep image prior (DIP) \cite{ulyanov2018deep} and Noise2Void \cite{krull2019noise2void}, and RGB images denoising: Restormer \cite{zamir2022restormer}, TECDNet \cite{zhao2022hybrid} and NAFnet \cite{chen2022simple}. %
The published weights of the compared methods performed poorly on our validation data, as our captured samples are severely noisy and exhibit different noise statistics due to variations in camera characteristics. To enable better comparison with our fine-tuned models, we fine-tuned these methods on the real-world training dataset we captured. Since the diffusion model predicts images from the posterior distribution, we averaged 20 predictions for each test sample to estimate the posterior distribution mean, ensuring a fair evaluation of the PSNR. 

\renewcommand{\noisy}[1]{figs/s21/noisy/sample0#1_low_res.png}
\renewcommand{\cycle}[1]{figs/s21/cycleisp_fix/0000#1_raw4c.pt.png}
\newcommand{\dip}[1]{figs/s21/dip/0000#1_raw4c.pt}
\renewcommand{\ntov}[1]{figs/s21/n2v/sample0#1.png}
\renewcommand{\concat}[1]{figs/s21/loraconcat/sample0#1.png}
\renewcommand{\cond}[1]{figs/s21/loracond/sample0#1.png}
\renewcommand{\gt}[1]{figs/s21/gt/gt0#1.png}
\renewcommand{\restormer}[1]{figs/s21/restormer/sample0#1.png}
\renewcommand{\NAFnet}[1]{figs/s21/NAFnet/sample0#1.png}

\renewcommand{\imgA}{16} 
\renewcommand{\imgB}{00} 
\renewcommand{\imgC}{04} 
\newcommand{\imgD}{03} 
\newcommand{\imgE}{17} 

\renewcommand{\pfig}[1]{\includegraphics[width=0.123\linewidth]{#1}}

\begin{figure*}[t]
    \setlength\tabcolsep{0.5pt} %
    \centering
    \begin{tabular}{cccccccc}
       
        Noisy S21 & CycleISP & Noise2Void & Restormer & NAFnet & Diffusion & Diffusion + & Clean S21  \vspace{-0.1cm}\\
        \footnotesize (input) & \footnotesize (raw) &\footnotesize (raw) &\footnotesize  (RGB) & \footnotesize 
 (RGB) & \footnotesize (raw , our) & Text \footnotesize (raw , our) & \footnotesize (target) \vspace{-0.1cm}\\
        \midrule
        \pfig{\noisy\imgA} & \pfig{\cycle\imgA} &  
        \pfig{\ntov\imgA} &
        \pfig{\restormer\imgA} &
        \pfig{\NAFnet\imgA} &
        \pfig{\concat\imgA}& \pfig{\cond\imgA} &
        \pfig{\gt\imgA}\\

        \pfig{\noisy\imgC} & \pfig{\cycle\imgC} &  
        \pfig{\ntov\imgC} &
        \pfig{\restormer\imgC} &
        \pfig{\NAFnet\imgC} &
        \pfig{\concat\imgC}& \pfig{\cond\imgC} &
        \pfig{\gt\imgC}\\
        
        \pfig{\noisy{\imgD}} & \pfig{\cycle{\imgD}} &  
        \pfig{\ntov{\imgD}} &
        \pfig{\restormer{\imgD}} &
        \pfig{\NAFnet{\imgD}} &
        \pfig{\concat{\imgD}}& \pfig{\cond{\imgD}} &
        \pfig{\gt{\imgD}}\\

        \pfig{\noisy{\imgE}} & \pfig{\cycle{\imgE}} &  
        \pfig{\ntov{\imgE}} &
        \pfig{\restormer{\imgE}} &
        \pfig{\NAFnet{\imgE}} &
        \pfig{\concat{\imgE}}& \pfig{\cond{\imgE}} &
        \pfig{\gt{\imgE}}\\

    \end{tabular}
    \caption{Real-world denoising comparison of various methods applied to Samsung S21 camera captures. Samples from the COCO dataset were displayed on a screen and captured twice under different settings to generate noisy and GT pairs. Our models were fine-tuned on real data using the proposed approach. Our text-guided model achieves superior results compared to competing methods, including a non-text-guided diffusion model. The tag raw/RGB indicates whether denoising was applied to raw or RGB images.}
    \label{fig:img_resultS21}
\end{figure*}

\begin{table}[t]
\caption{Quantitative comparison of results for simulated noise on COCO dataset samples. Two noise levels are presented: low (0.1) and high (0.3). Our method performs better in perceptual metrics (LPIPS, DISTS) rather than in PSNR, according to the characteristics of our method (\Cref{sec:results}).}
  \centering
  \setlength\tabcolsep{2pt}
  
\resizebox{0.9\linewidth}{!}{
  \begin{tabular}{lccccc}
    \toprule
    Method & \multicolumn{2}{c}{PSNR $\uparrow$} & SSIM$\uparrow$ & LPIPS$\downarrow$ & DISTS$\downarrow$ \vspace{-3pt}\\
     & {\scriptsize RAW} & {\scriptsize RGB} \vspace{-3pt}\\
    \midrule
     \multicolumn{6}{@{}c}{\textbf{\textit{Low noise (0.1)}}}\\
    CycleISP \cite{zamir2020cycleisp} & {26.84} &21.59 & 0.589 & 0.394 & 0.270\\
    DIP \cite{ulyanov2018deep}& 22.00 & 13.83 & 0.413 & 0.632 & 0.448\\
    Noise2Void \cite{krull2019noise2void} & {26.52} &20.45 & 0.614 & 0.418 & 0.285\\

    Diffusion {\scriptsize(our)}  & \underline{27.93} &\textbf{24.10} & \textbf{0.640} & \underline{0.255} & \underline{0.190}\\
    + Text  {\scriptsize (our)} & \textbf{29.72} &\underline{24.00} & \underline{0.629} & \textbf{0.250} & \textbf{0.182} \\
    \midrule
         \multicolumn{6}{@{}c}{\textbf{\textit{High noise (0.3)}}}\\

    CycleISP \cite{zamir2020cycleisp} & {24.25}&19.18 & 0.440 & 0.537 & 0.322  \\
    Noise2Void \cite{krull2019noise2void}& {25.14} &{20.02} & \textbf{0.567} & 0.479 & 0.309  \\

    Diffusion {\footnotesize(our)} & \underline{25.23} &\underline{21.73} & 0.515 & \underline{0.411} & \underline{0.243}  \\
    + Text  {\footnotesize(our)} & \textbf{25.24} &\textbf{21.78} & \underline{0.516} & \textbf{0.397} & \textbf{0.228}  \\
    \bottomrule
  \end{tabular}}
\vspace{-0.1in}    
\label{tab:result_sim0103}
\end{table}

\noindent \textbf{Simulated Noise.}
We evaluated our approach using two levels of simulated noise. The lower noise level, denoted as '0.1', was generated using our noise model (\Cref{eq:noise_lambda_params_sample}) with parameters $\log\lambda_{shot}=0.1$ and $\log\lambda_{read}=0.2$. The higher noise level, denoted as '0.3', was generated with $\log\lambda_{shot}=0.3$ and $\log\lambda_{read}=0.5$. The synthetic noise was sampled with a fixed random seed for all tests, ensuring consistency across evaluations. Competing methods were trained on the same dataset and noise simulation parameters.

Quantitative results for the two noise levels (0.1 and 0.3) are presented in \Cref{tab:result_sim0103}. As shown in the table, the diffusion method outperforms other approaches in both structural and perceptual metrics, specifically LPIPS and DISTS. Visual differences are also clearly observable in \Cref{fig:sim_result01} and the supplementary material. However, in terms of PSNR and SSIM, the results are less conclusive due to the nature of the diffusion method, which prioritizes perceptual quality.

For the text-guided model's sampling process, we used the original captions from the COCO dataset for both simulated noise and real-world captures. The contribution of text guidance can be evaluated by comparing our non-text-guided diffusion model to the text-guided model. Since both models were trained under similar conditions, the only distinction is the inclusion of scene captions. This information about the scene's content enhances reconstructing textures and fine details that the non-text-guided model is unable to achieve.

\noindent \textbf{Real World Noise.} %
In the second stage of our approach, the gap between simulated noise and the target camera sensor noise (Samsung-S21 or Allied Vision G-146C) is bridged through low-rank fine-tuning of the diffusion model, as detailed in \Cref{sec:method}.
Testing our method on real sensor images with true noise achieved superior performance compared to prior state-of-the-art raw and RGB denoisers. Results for the Samsung-S21 camera are presented in \Cref{fig:img_resultS21,tab:resultLORA} and those for the Allied Vision camera are provided in \Cref{tab:resultAllied} and the supplementary material. Given the strong performance of all the tested models on the Allied Vision images, we conclude that this dataset is less noisy and easier to process compared to the Samsung-S21 data. In such low-noise images, the denoising task becomes less ill-posed, and the contribution of the additional text information is less pronounced. Overall, our method outperforms others in all image metrics as can be seen in \Cref{tab:resultAllied}. 

Since the diffusion model maximizes perceptual quality by sampling from the posterior distribution rather than minimizing mean square error, our method does not outperform others in terms of SSIM and PSNR. This outcome aligns with the `perception-distortion tradeoff' discussed in \Cref{sec:relatedwork}, as well as in \cite{yu2024scaling}. \Cref{fig:img_resultS21} visually shows the improved results.  Additionally, the CLIP score (CS) measures the correspondence between the text caption and the image. We used the latest CLIP version (ViT-L/14@336px) for the most accurate evaluation, computing the cosine similarity between the embeddings of the resulting image and the text caption. Notably,  the text-guided model results are more consistent with the image’s textual description. 
The contribution of the text caption is prominent compared to the diffusion model without text input, as shown in \Cref{fig:img_resultS21,fig:teaser}.
\Cref{fig:img_resultS21_realcam} presents additional results of outdoor scenes and real-world captures (without corresponding reference images). Our proposed method demonstrates superior reconstruction results for real images captured outside the lab across varied scenes and under different lighting conditions. %

\begin{table}[t]
\caption{Quantitative comparison of real sensor noise results using the Samsung-S21 smartphone camera. Our method achieves the highest performance in PSNR (both raw and RGB) and perceptual metrics. Additionally, the CLIP score (CS), which quantifies text-image similarity, highlights the superior performance of our text-guided model.}

  \centering
  \setlength\tabcolsep{2pt}
  
\resizebox{1.03\linewidth}{!}{%
  \begin{tabular}{lcccccc}
    \toprule
    Method & \multicolumn{2}{c}{PSNR $\uparrow$} & SSIM$\uparrow$ & LPIPS$\downarrow$ & DISTS$\downarrow$ & CS$\uparrow$ \vspace{-3pt}\\
    & {\scriptsize RAW} & {\scriptsize RGB} \vspace{-3pt} \\ 
    \midrule
    CycleISP {\footnotesize paper} & 20.95 &14.59 & 0.111 & 0.862 & 0.506 & 20.50\\
    CycleISP {\footnotesize trained} & 24.50 & 21.75 & {0.618} & 0.445 & 0.288 & \underline{22.69}\\

    DIP & 23.60 & 18.41 & 0.460 & 0.629 & 0.360 & 19.81\\
    Noise2Void & {24.25} & 20.52 & 0.574 & 0.488 & 0.306 & 21.38\\
    Restormer \scriptsize (RGB) &
    -- & 21.94 &
    \textbf{0.635} & 0.427 & 0.288 & 21.94\\
    NAFnet \scriptsize (RGB) & 
    -- & \underline{21.97} &
    {0.627} & 0.423 & 0.276 & {22.42}\\
    TECDNet \scriptsize (RGB) & 
    -- & \textbf{22.07} &
    \textbf{0.636} & 0.427 & 0.289 & {22.51}\\

    Diffusion \scriptsize (our) & \textbf{24.60} & \textbf{22.07} & {0.583} & \underline{0.322} & \underline{0.219} & {21.93}\\
    + Text \scriptsize (our) & \underline{24.57}& 21.95 & {0.589} & \textbf{0.300} & \textbf{0.204} & \textbf{23.34}\\

    \bottomrule
  \end{tabular}}
      \label{tab:resultLORA}
\end{table}

\renewcommand{\noisy}[1]{figs/realcam/noisy/sample0#1_low_res.png}
\newcommand{\cyclepaper}[1]{figs/realcam/cycleisp_paper/#1.png}
\renewcommand{\cycle}[1]{figs/realcam/cycleisp_trained/#1.png}

\renewcommand{\ntov}[1]{figs/realcam/n2v/sample0#1.png}
\renewcommand{\restormer}[1]{figs/realcam/restormer/sample0#1.png}
\renewcommand{\NAFnet}[1]{figs/realcam/NAFnet/sample0#1.png}

\renewcommand{\concat}[1]{figs/realcam/concat/sample0#1.png}
\renewcommand{\cond}[1]{figs/realcam/cond/sample0#1.png}

\renewcommand{\imgA}{08} 
\renewcommand{\imgB}{17} 
\renewcommand{\imgC}{11} %
\renewcommand{\imgD}{32} 
\renewcommand{\imgE}{20000}  %
\renewcommand{\pfig}[1]{\raisebox{-1.8cm}{\includegraphics[width=0.13\linewidth]{#1}}}
\renewcommand{\pfig}[1]{\includegraphics[width=0.12\linewidth]{#1}}

\begin{figure*}[!t]
    \setlength\tabcolsep{1pt} %
    \centering
    \begin{tabular}{cccccc|cc}
       
        \multicolumn{1}{c|}{Noisy S21} & CycleISP   & CycleISP  & Noise2Void & Restormer & NAFnet & Diffusion & Diffusion +  \vspace{-0.1cm}\\ 
        \multicolumn{1}{c|}{ } & \footnotesize (raw , paper)  &
        \footnotesize(raw) & \footnotesize(raw) 
        &\footnotesize (RGB)& \footnotesize (RGB)& \footnotesize (raw , our) &  Text \footnotesize (raw , our)  \vspace{-0.1cm}\\

        \midrule
        \multicolumn{6}{l}{caption: A blue road sign with two white arrows and a white sign with text.} \\

        \pfig{\noisy\imgA} & \pfig{\cyclepaper\imgA} &  \pfig{\cycle\imgA} &
        \pfig{\ntov\imgA} &
        \pfig{\restormer\imgA} &
        \pfig{\NAFnet\imgA} &

        \pfig{\concat\imgA}&  \pfig{\cond\imgA} \\ %
        \multicolumn{6}{l}{caption: A car on a street with a tree.} \\
        \pfig{\noisy\imgB} & \pfig{\cyclepaper\imgB} &  \pfig{\cycle\imgB} & \pfig{\ntov\imgB} &
        \pfig{\restormer\imgB} &
        \pfig{\NAFnet\imgB} &
        \pfig{\concat\imgB} &  \pfig{\cond\imgB} \\ %
        \multicolumn{6}{l}{caption: A red scooter on the street.} \\
        \pfig{\noisy\imgC} & \pfig{\cyclepaper\imgC} &  \pfig{\cycle\imgC} &
        \pfig{\ntov\imgC} &
        \pfig{\restormer\imgC} &
        \pfig{\NAFnet\imgC} &
        
        \pfig{\concat\imgC} & \pfig{\cond\imgC} \\ %
        \multicolumn{6}{l}{caption: Red and white curbstone with grass and road.} \\
        \pfig{\noisy{\imgD}} & \pfig{\cyclepaper{\imgD}} &  \pfig{\cycle{\imgD}} & \pfig{\ntov{\imgD}} &        
        \pfig{\restormer\imgD} &
        \pfig{\NAFnet\imgD} &
        \pfig{\concat{\imgD}} & \pfig{\cond{\imgD}} \\ %

    \end{tabular}
    \caption{Outdoor real-world reconstruction results for images captured using a Samsung S21 smartphone camera under low-light conditions. Since these captures were performed outside the controlled environment of the lab, no ground truth (GT) images are available. Our models were fine-tuned on real data following our proposed approach. The tag raw/RGB indicates whether denoising was applied to raw or RGB images.}
    \label{fig:img_resultS21_realcam}
\end{figure*}

\noindent \textbf{Data Contribution.} 
We collected a new dataset containing raw images of captionable objects with descriptive captions. %
We utilized screen capturing to capture samples from the COCO-Captions dataset, capturing pairs of noisy and clean images using a static camera and different camera settings. This approach retained the relevance of the original captions of the COCO dataset to the captured images.

For the Samsung S21 camera, the noisy images were captured with an exposure time of $1/12000$ s and ISO 3200, while the clean target images were captured with an exposure time of $1/50$ s and ISO 50. For the Allied Vision camera, the noisy images used an exposure time of $175\mu s$ and a gain of 32.04, whereas the clean images were captured with an exposure time of $7 ms$ and zero gain.

\noindent \textbf{Ablation and Limitations} %
Fine-tuning the model on the real-world dataset (\Cref{tab:resultLORA}) improved the results in all perceptual metrics compared to the models trained only on simulated noise ("base model" in \Cref{tab:ablation}). Thus, utilizing LoRA with a small amount of data enables closing the gap between the different noise statistics. \Cref{tab:ablation} shows that we may add text to the base model to improve results also without the LoRA fine-tuning. Clearly, using LoRA improves even further (\Cref{tab:resultLORA}).

The diffusion model reconstructs missing features and details in the noisy image based on the learned image prior, conditioned on the input text. Since sensor noise destructively remove true content, the generated features may not perfectly match the original scene. However, perceptual metrics indicate that they remain visually similar. To assess the contribution of the text, we tested our text-guided model with irrelevant (i.e. wrong) captions during the reconstruction (\Cref{tab:ablation} and in the supplementary). It resulted in a drop in all perceptual metrics, especially noticeable in the CLIP score, which measures the alignment of the results with the true captions. This performance decrease occurs because while many possible clean images correspond to the noisy input, the diffusion estimates a sample that also aligns with the input text, and while the text is incorrect, the result features are less accurate. The visual influence of the text is presented in \Cref{fig:text_ablation}. While the overall appearance of the image stays unchanged, fine details are reconstructed according to the text description: "fluffy" and "far" generate far texture, "smooth hard plastic" generates smoother texture and the irrelevant caption as "person on the street" produces features that can be assumed to be suitable for the street, such as stone wall and straight lines.
This example shows that the method is unlikely to produce images that are not real or fabricated since the solution space is limited to images that satisfy the noisy input structure while the text assists in completing the missing features and details. 
\begin{table}[t]
\caption{\textbf{Ablation study on real sensor images:} (1) Results of our method before LORA fine-tuning on real-world data, highlighting the contribution of fine-tuning, and (2) results of the proposed model using irrelevant (wrong) captions, demonstrating the significance of captions in the process. All results are compared to our best results in \Cref{tab:resultLORA}.}

\label{tab:ablation}
  \centering
  \setlength\tabcolsep{2pt}
    \vspace{-0.04in}
  
\resizebox{1.02\linewidth}{!}{%
  \begin{tabular}{lcccccc}
    \toprule
    Method & \multicolumn{2}{c}{PSNR $\uparrow$} &  SSIM$\uparrow$ &  LPIPS$\downarrow$ &  DISTS$\downarrow$ &  CS$\uparrow$ \vspace{-3pt}\\
    & {\scriptsize RAW} & {\scriptsize RGB} \vspace{-3pt} \\ 
    \midrule
    \multicolumn{6}{@{}l}{(1) \textbf{Base model} (without LORA)} \\
    Diffusion & 24.35 & 22.10 & {0.536} & {0.442} & {0.272} & 19.56\\
    + Text & 24.34&  22.63 & {0.566} & {0.408} & {0.251} & 22.32\\
    \multicolumn{6}{@{}l}{(2) \textbf{With LORA} (compare to \Cref{tab:resultLORA})} \\
    irrelevant caps. & 24.15 &(21.21) & 0.584 & 0.310 & 0.219 & 21.6\\
    \bottomrule
  \end{tabular}}
\end{table}

The reliance on the quality and relevance of the captions may reveal a limitation of our approach as it means that scenes with ambiguous or complex details may not be accurately reconstructed. Another limitation is the high computational cost and complexity of training and deploying diffusion models, which may limit their accessibility and scalability for real-time or resource-constrained applications. This may be mitigated by using recent methods to improve diffusion efficiency such as \cite{Song2023Consistency,Abu-Hussein2024UDPM}.

\renewcommand{\pfig}[1]{\raisebox{-.5\totalheight}{\includegraphics[width=0.33\linewidth]{#1}}}

\begin{figure}[t]
    \setlength\tabcolsep{1pt} %
    \centering
    \begin{tabular}{ccc}

        \pfig{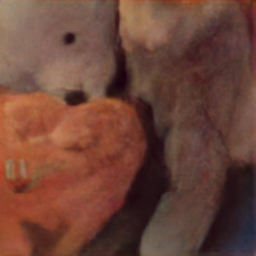} &
        \pfig{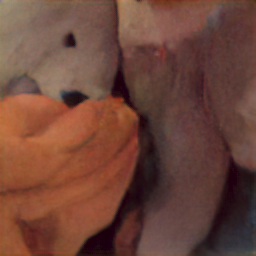} &
        \pfig{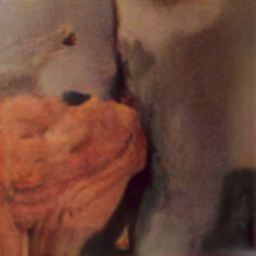}  \vspace{2mm} \\

        \small \shortstack{A soft and fluffy \\ doll fur} &
        \small \shortstack{A smooth hard \\ plastic doll} &
        \small \shortstack{A person on \\ the street} \\
        
    \end{tabular}
    \caption{Different text inputs influence the textures and fine details of the generated image, while the original structure of the image is maintained due to the dependence on the input image data.}
    \label{fig:text_ablation}
    \vspace{-0.1in}
\end{figure}

\begin{table}[t]

\caption{Quantitative comparison of real sensor noise results using the Allied Vision Manta G-146C camera. This dataset exhibits lower noise levels, making the denoising problem less ill-posed, and all tested models achieve relatively strong results. Nonetheless, our approach demonstrates superior perceptual quality compared to other methods
}

  \centering
  \setlength\tabcolsep{2pt}
  
\resizebox{1.03\linewidth}{!}{%
  \begin{tabular}{lcccccc}
    \toprule
    Method & \multicolumn{2}{c}{PSNR $\uparrow$} & SSIM$\uparrow$ & LPIPS$\downarrow$ & DISTS$\downarrow$ & CS$\uparrow$ \vspace{-3pt}\\
    & {\scriptsize RAW} & {\scriptsize RGB} \vspace{-3pt} \\ 
    \midrule
    CycleISP {\footnotesize paper} & 30.71 &20.44 & 0.764 & 0.317 & 0.234 & 24.63\\
    CycleISP {\footnotesize trained} & 32.47 & 24.53 & 0.761 & 0.245 & 0.195 & \underline{25.89}\\

    Restormer \scriptsize (RGB) &
    -- & 25.30 &
    0.797 & \underline{0.184} & 0.171 & 25.04\\
    NAFnet \scriptsize (RGB) & 
    -- & 26.22 & 0.799 & 0.186 & 0.173 & 25.55\\
    TECDNet \scriptsize (RGB) & 
    -- & 25.18 & 0.795 & 0.227 & 0.192 & \textbf{25.91}\\

    Diffusion \scriptsize (our) & \textbf{33.61}	& \textbf{26.53} & \textbf{0.808}	& \textbf{0.163}	& \underline{0.145} & {25.38}\\
    + Text \scriptsize (our) & \underline{33.48}& \underline{26.26} & \underline{0.806} & \textbf{0.163} & \textbf{0.144} & {25.85}\\

    \bottomrule
  \end{tabular}}
      \label{tab:resultAllied}
      
\end{table}

\section{Conclusion}
\label{sec:conclusion}

This work presented a novel text-guided approach to real-world noisy image enhancement, demonstrating significantly improved reconstruction compared to non-text-guided models. These promising results suggest a new paradigm in photography that integrates human vision, specifically the photographer’s input, through captions with advanced image processing and enhancement algorithms. By incorporating descriptive information, our method may push the boundaries of current imaging techniques, offering a more interactive and intuitive approach to achieve high-quality image reconstruction, even under challenging conditions.

{
\bibliographystyle{ieee_fullname}
\bibliography{mybib}
}
\clearpage
\appendices

In this supplementary material, we provide additional details and results for the paper \textit{Tell Me What You See: Text-Guided Real-World Image Denoising}. In the following sections, we describe our captured dataset setup and the parameters used, we provide further details on our methodology, and present additional results for both simulated and real-world noise.

\section{Data Contribution} 
For text-guided real-world image denoising, we require raw noisy images paired with corresponding text captions that describe the scene. Existing denoising datasets primarily consist of image crops, where each crop contains only a specific pattern or structure, lacking captionable content or identifiable objects.

To address this, we collected a new dataset tailored to our use case, containing images of captionable objects, each has a descriptive caption of the scene. The noisy images were saved in raw format to ensure an effective and accurate denoising process, as opposed to working with RGB images. We utilized screen capturing to capture samples from the COCO-Captions dataset, capturing pairs of noisy and clean images using a static camera and different camera settings. This approach retained the relevance of the original captions of the COCO dataset to the captured images.

We disregarded artifacts that might arise from screen capturing, such as Moiré patterns and color variations since the ground truth images were also captured using the same method. The primary difference between the clean and noisy images is the noise levels, which result from variations in camera settings and parameters.

For the Samsung S21 camera, the noisy images were captured with an exposure time of $1/12000$ s and ISO 3200, while the clean target images were captured with an exposure time of $1/50$ s and ISO 50. For the Allied Vision camera, the noisy images used an exposure time of $175\mu s$ and a gain of 32.04, whereas the clean images were captured with an exposure time of $7 ms$ and zero gain.

All datasets, along with corresponding captions and metadata, will be published to support further research and benefit the research community.

\definecolor{cy}{rgb}{1, 0.95, 0.8}%
\definecolor{cb}{rgb}{0.92, 0.94, 1} 
\renewcommand{\hlc}[2][yellow]{{%
    \colorlet{foo}{#1}%
    \sethlcolor{foo}\hl{#2}}}
\begin{figure}[tb] %
    \centering
    \vspace{-0.001in}
    \includegraphics[width=\linewidth]{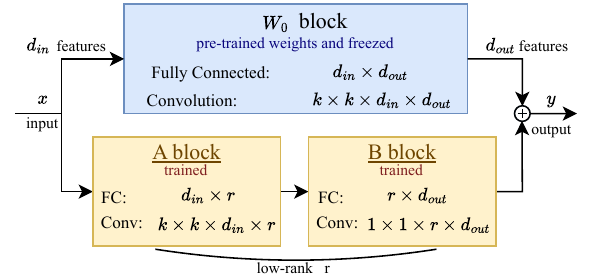}
    \caption{Details of low-rank adaptation (LoRA) used for model fine-tuning. Fully connected and convolutional layers were adapted by adding trained low-rank bypass layers $A$ and $B$ (in \hlc[cy!120]{\textbf{yellow}}) while keeping the pre-trained model weights fixed (in \hlc[cb!150]{\textbf{blue}}).
    For fully connected layers the dimensions of the weights are formatted as \texttt{in\_featurs x out\_features}. For convolutional layers, the dimensions of the weights are formatted as \texttt{k\_h x k\_w x in\_features x out\_features}, where $k_h \times k_w$ denotes the kernel size.
    }

    \label{fig:lora_diagram}
\end{figure}

\renewcommand{\pfig}[1]{\includegraphics[width=0.35\linewidth]{#1}}
\begin{figure}[t]
    \begin{tabular}{ccc}
    \rotatebox{90}{\parbox[c]{3cm}{\centering Noisy }} & \pfig{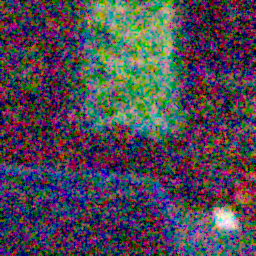} & 
        \pfig{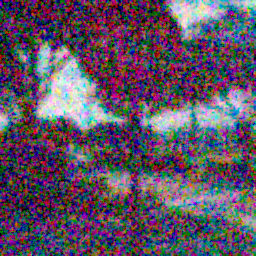} \\
    \midrule
    \rotatebox{90}{\parbox[c]{2cm}{\centering Wrong text }} & \pfig{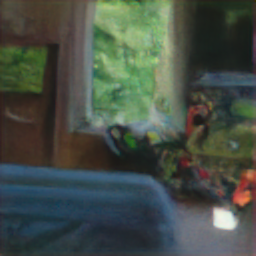} & 
        \pfig{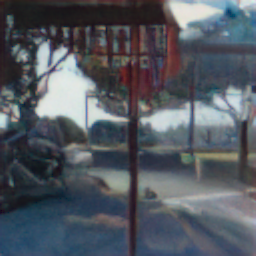} \\
    & \textit{\shortstack{A person on a motor \\bike travels around \\a sharp corner.}} &
    \textit{\shortstack{A narrow hotel room with \\ two made up beds.}}\\

    \midrule
    \rotatebox{90}{\parbox[c]{2.3cm}{\centering Correct text }} & \pfig{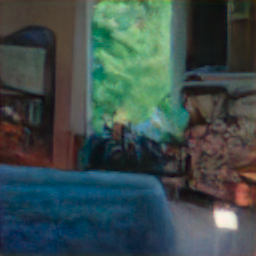} & 
        \pfig{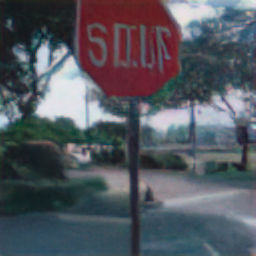} \\
    & \textit{\shortstack{Bedroom scene with \\a bookcase, blue \\comforter and window.}} &
    \textit{\shortstack{A stop sign is mounted \\ upside-down on it's post.}}\\
    \midrule
    \rotatebox{90}{\parbox[c]{2.3cm}{\centering GT }} & \pfig{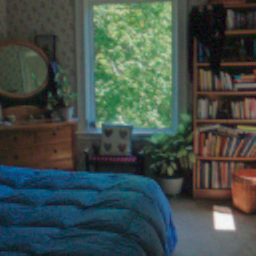} & 
        \pfig{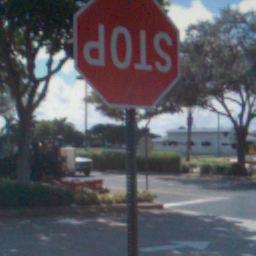} \\

    \end{tabular}
    \caption{Comparison of the text-guided diffusion model results using incorrect text versus the correct text. Incorrect captions lead to performance degradation due to inaccurate reconstruction of details.}
    \label{fig:wrongtext_ablation}
\end{figure}

\section{COCO captions - test samples}\label{supp:cocodata}
We used the captions provided in the COCO-Captions dataset \cite{chen2015microsoft} for image reconstruction with our text-guided model. These data samples were used in our experiments and evaluations involving both simulated noise and real-world noise captured by our cameras. 
Several ground truth samples from the dataset are shown in \Cref{fig:coco_cap_data}. For quantitative evaluations, we used larger test set of samples than those presented here.

\section{LoRA Fine-Tuning Details}
The LoRA \cite{hu2021lora} method adds a low-rank set of trainable weights to a pre-trained model for fine-tuning. The original model remains fixed while the low-rank weights are optimized on a smaller dataset. As described in the main paper, these low-rank weights were added to the fully connected layers of the attention modules and the convolutional layers of the residual blocks. The architecture details are illustrated in \Cref{fig:lora_diagram}. The dimensions of the fully connected layer weights are given by \verb|in_featurs x out_features|, while for the convolutional layers the dimensions are \verb|k_h x k_w x in_features x out_features|, where $k_h \times k_w$ denotes the kernel size.

\section{Limitations - Wrong Text Results}
As discussed in the main paper, incorrect text input leads to inaccurate reconstructions and a drop in performance. \Cref{fig:wrongtext_ablation} illustrates the performance degradation caused by incorrect text descriptions, highlighting the sensitivity of our method to text input and its impact on reconstruction quality. The diffusion model attempts to generate an image that aligns with the provided text, and incorrect text results in inaccurate details and features in the image.

\section{Additional Results}
Additional results for low (0.1) and high (0.3) simulated noise levels are presented in \Cref{fig:img_result01_more} and \cref{fig:sim_result03B}, respectively. \Cref{fig:LORA} demonstrates the improvement achieved through LoRA fine-tuning compared to the base model, which was trained solely on simulation data. Real-world noise results from Samsung S21 captures are shown in \Cref{fig:img_resultS21_more2}, while results for Allied Vision captures are provided in \Cref{fig:img_resultSallied}.

\def\gt#1{figs/coco_sim/gt/gt0#1.png}
\def\pfig#1{\raisebox{-1cm}{\includegraphics[width=0.1\linewidth]{#1}}}

\begin{table*}
    \caption{Examples of COCO Captions data samples used for testing: images and text captions.}
    \centering
    \begin{tabular}{l|p{5cm}|l|p{5cm}}
       
        Image & Text caption & Image & Text caption  \\
        \midrule
        \vspace{2pt}
        \pfig{\gt{00}} & A room with chairs, a table, and a woman in it. &\vspace{2pt}\pfig{\gt{04}} & Three teddy bears, each a different color, snuggling together. \\
        \vspace{2pt}\pfig{\gt{01}} & A big burly grizzly bear is show with grass in the background. &\vspace{2pt}\pfig{\gt{05}} &  A woman posing for the camera standing on skis.\\
        \vspace{2pt}\pfig{\gt{02}} & Bedroom scene with a bookcase, blue comforter and window. &\vspace{2pt}\pfig{\gt{16}} & A street scene with focus on the street signs on an overpass. \\
        \vspace{2pt}\pfig{\gt{03}} & A stop sign is mounted upside-down on it's post. & \vspace{4pt}\pfig{\gt{17}} & A red double decker bus driving down a city street. \\

    \end{tabular}
    \label{fig:coco_cap_data}
\end{table*}

\def\cycle{figs/coco_sim/cycleisp01/fix}
\def\ntov{figs/coco_sim/n2v}
\def\noisy{figs/coco_sim/noisy}
\def\concat{figs/coco_sim/concat}
\def\cond{figs/coco_sim/cond}
\def\gt{figs/coco_sim/gt}

\def\imgA{01} 
\def\imgB{03} 
\def\imgC{05} 
\def\imgD{16} 
\def\imgE{17} 

\def\imnum#1{0000#1.pt.png} 
\def\dimnum#1{sample0#1.png}
\def\gtimnum#1{gt0#1.png}

\def\nlvl{01}

\renewcommand{\pfig}[1]{\includegraphics[width=0.14\linewidth]{#1}}

\begin{figure*}[!ht]
    \setlength\tabcolsep{4pt} %
    \centering
    \begin{tabular}{cccccc}
        Noisy 0.1 & CycleISP & Noise2Void & Diffusion & Diffusion & GT 
        \vspace{-0.1cm}\\
        & & & \footnotesize (our) &  +Text \footnotesize (our)  & \vspace{-0.1cm}\\
        \midrule
        \pfig{\noisy\nlvl/\imnum\imgA} & \pfig{\cycle\nlvl/\imnum\imgA} & 
        \pfig{\ntov\nlvl/\dimnum\imgA} &
        \pfig{\concat\nlvl/\dimnum\imgA}& \pfig{\cond\nlvl/\dimnum\imgA} &
        \pfig{\gt/\gtimnum\imgA}\\
    
        \pfig{\noisy\nlvl/\imnum\imgB} & \pfig{\cycle\nlvl/\imnum\imgB} & 
        \pfig{\ntov\nlvl/\dimnum\imgB}&
        \pfig{\concat\nlvl/\dimnum\imgB}& \pfig{\cond\nlvl/\dimnum\imgB} &
        \pfig{\gt/\gtimnum\imgB} \\
    
        \pfig{\noisy\nlvl/\imnum\imgC} & \pfig{\cycle\nlvl/\imnum\imgC} & 
        \pfig{\ntov\nlvl/\dimnum\imgC}&
        \pfig{\concat\nlvl/\dimnum\imgC}& \pfig{\cond\nlvl/\dimnum\imgC}&
        \pfig{\gt/\gtimnum\imgC}\\

        \pfig{\noisy\nlvl/\imnum\imgD} & \pfig{\cycle\nlvl/\imnum\imgD} & 
        \pfig{\ntov\nlvl/\dimnum\imgD}&
        \pfig{\concat\nlvl/\dimnum\imgD}& \pfig{\cond\nlvl/\dimnum\imgD}&
        \pfig{\gt/\gtimnum\imgD}\\

        \pfig{\noisy\nlvl/\imnum\imgE} & \pfig{\cycle\nlvl/\imnum\imgE} & 
        \pfig{\ntov\nlvl/\dimnum\imgE}&
        \pfig{\concat\nlvl/\dimnum\imgE}& \pfig{\cond\nlvl/\dimnum\imgE}&
        \pfig{\gt/\gtimnum\imgE}\\
        
    \end{tabular}
    \caption{\textbf{Low simulated noise results.} Comparison of various methods for raw image denoising at a noise level of 0.1 ($\log\lambda_{shot}=0.1$ and  $\log\lambda_{read}=0.2$). Our models achieve superior results while the text guidance contributes to the results' perceptual quality, details, and textures.}
    \label{fig:img_result01_more}
\end{figure*}

\def\cycle{figs/coco_sim/cycleisp01/fix}
\def\ntov{figs/coco_sim/n2v}
\def\noisy{figs/coco_sim/noisy}
\def\concat{figs/coco_sim/concat}
\def\cond{figs/coco_sim/cond}
\def\gt{figs/coco_sim/gt}
\def\NAFnet{figs/coco_sim/NAFnet}
\def\restormer{figs/coco_sim/restormer}

\def\imgA{0}
\def\imgB{2}
\def\imgC{4}

\def\imnum#1{00000#1.pt.png}
\def\dimnum#1{sample00#1.png}
\def\gtimnum#1{gt00#1.png}

\def\imnumA{000003.pt.png}
\def\dimnumA{sample003.png}

\def\imnumB{000004.pt.png}
\def\dimnumB{sample004.png}
\def\imnumC{000005.pt.png}
\def\dimnumC{sample005.png}
\def\nlvl{01}

\def\pfig#1{\includegraphics[width=0.13\linewidth]{#1}}

\renewcommand{\cycle}{figs/coco_sim/cycleisp03/fix}
\renewcommand{\imgA}{5} 
\renewcommand{\imgB}{1} 
\renewcommand{\imgC}{3} 
\renewcommand{\nlvl}{03}
\renewcommand{\pfig}[1]{\raisebox{-.5\totalheight}{\includegraphics[width=0.13\linewidth]{#1}}}

\begin{figure*}[!t]
    \setlength\tabcolsep{2pt} %

    \centering
    \begin{tabular}{cccccm{1.9cm}c}
        Noisy 0.3 & CycleISP & Noise2Void &  Diffusion &\multicolumn{2}{|c|}{Diffusion + Text} &\hspace{0.5mm} GT \vspace{-0.1cm}\\
        & & & \footnotesize (our) &  \multicolumn{2}{|c|}{\footnotesize (our)}  & \vspace{-0.1cm}\\

        \midrule
        \pfig{\noisy\nlvl/\imnum\imgA} & \pfig{\cycle\nlvl/\imnum\imgA} & 
        \pfig{\ntov\nlvl/\dimnum\imgA} &
                
        \pfig{\concat\nlvl/\dimnum\imgA}& \pfig{\cond\nlvl/\dimnum\imgA} &
        {\small A woman posing for the camera standing on skis.} &\hspace{0.5mm}
        \pfig{\gt/\gtimnum\imgA}\\
    
        \pfig{\noisy\nlvl/\imnum\imgB} & \pfig{\cycle\nlvl/\imnum\imgB} & 
        \pfig{\ntov\nlvl/\dimnum\imgB}&
        
        \pfig{\concat\nlvl/\dimnum\imgB}& \pfig{\cond\nlvl/\dimnum\imgB} &
        \small A big burly grizzly bear is show with grass in the background. &\hspace{0.5mm}
        \pfig{\gt/\gtimnum\imgB} \\
    
        \pfig{\noisy\nlvl/\imnum\imgC} & \pfig{\cycle\nlvl/\imnum\imgC} & 
        \pfig{\ntov\nlvl/\dimnum\imgC}&
         
        \pfig{\concat\nlvl/\dimnum\imgC}& \pfig{\cond\nlvl/\dimnum\imgC}&
        \small A stop sign is mounted upside-down on it’s post.. &\hspace{0.5mm}
        \pfig{\gt/\gtimnum\imgC}\\

    \end{tabular}
    \caption{\textbf{High simulated noise results.} Comparison of various methods for raw image denoising at a noise level of 0.3 ($\log\lambda_{shot}=0.3$ and  $\log\lambda_{read}=0.5$).}
    \label{fig:sim_result03B}
\end{figure*}

\renewcommand{\noisy}[1]{figs/s21/noisy/sample0#1_low_res.png}
\renewcommand{\cycle}[1]{figs/s21/cycleisp_fix/0000#1_raw4c.pt.png}
\renewcommand{\ntov}[1]{figs/s21/n2v/sample0#1.png}
\renewcommand{\concat}[1]{figs/s21/loraconcat/sample0#1.png}
\renewcommand{\cond}[1]{figs/s21/loracond/sample0#1.png}
\renewcommand{\gt}[1]{figs/s21/gt/gt0#1.png}

\def\dip#1{figs/s21/dip/0000#1_raw4c.pt}

\newcommand{\concatA}[1]{figs/s21/baseconcat/sample0#1.png}
\newcommand{\condA}[1]{figs/s21/basecond/sample0#1.png}

\renewcommand{\pfig}[1]{\includegraphics[width=0.13\linewidth]{#1}}

\renewcommand{\imgA}{00} 
\renewcommand{\imgB}{05} 
\renewcommand{\imgC}{01} 
\renewcommand{\imgD}{03} 
\renewcommand{\imgE}{07} 
\begin{figure*}[p]
    \centering
    \begin{tabular}{c|cc|cc|c}
       Noisy S21 & \multicolumn{2}{c|}{Diffusion} & \multicolumn{2}{c|}{{Diffusion +Text}} & Clean S21 \\
         & Base model & + LoRA & Base model & + LoRA &  \\
        \midrule
        \pfig{\noisy\imgA} & \pfig{\concatA\imgA} &  \pfig{\concat\imgA} &
        \pfig{\condA\imgA}& \pfig{\cond\imgA} &
        \pfig{\gt\imgA}\\
    
        \pfig{\noisy\imgB} & \pfig{\concatA\imgB} &  \pfig{\concat\imgB} &
        \pfig{\condA\imgB}& \pfig{\cond\imgB} &
        \pfig{\gt\imgB}\\
    
        \pfig{\noisy\imgC} & \pfig{\concatA\imgC} &  \pfig{\concat\imgC} &
        \pfig{\condA\imgC}& \pfig{\cond\imgC} &
        \pfig{\gt\imgC}\\
        
        \pfig{\noisy\imgD} & \pfig{\concatA\imgD} &  \pfig{\concat\imgD} &
        \pfig{\condA\imgD}& \pfig{\cond\imgD} &
        \pfig{\gt\imgD}\\

    \end{tabular}
    \caption{\textbf{LoRA Fine-Tuning Improvement.} Fine-tuning the diffusion model with LoRA on real-world sensor noise enhances performance for both non-text-guided and text-guided models. This fine-tuning contributes to better generalization to real-world noise, as demonstrated qualitatively.}

    \label{fig:LORA}
\end{figure*}

\def\imnum#1{0000#1.pt.png} 
\def\dimnum#1{sample0#1.png}
\def\gtimnum#1{gt0#1.png}

\def\nlvl{01}

\renewcommand{\pfig}[1]{\includegraphics[width=0.14\linewidth]{#1}}

\renewcommand{\imgA}{00} 
\renewcommand{\imgB}{01} 
\renewcommand{\imgC}{02} 
\renewcommand{\imgD}{05} 
\renewcommand{\imgE}{06} 
\renewcommand{\pfig}[1]{\includegraphics[width=0.13\linewidth]{#1}}

\def\restormer#1{figs/s21/restormer/sample0#1.png}

\newcommand{\nafnet}[1]{figs/s21/NAFnet/sample0#1.png}
\renewcommand{\pfig}[1]{\includegraphics[width=0.128\linewidth]{#1}}

\begin{figure*}[p]
\vspace{-0.2in}
    \setlength\tabcolsep{5pt} %
    \renewcommand{\arraystretch}{0.86}
    \centering
    \begin{tabular}{cccccc}
        \rotatebox{90}{\parbox[c]{2.3cm}{\centering Noisy S21 \\ \footnotesize (input)}} & \pfig{\noisy\imgA} & \pfig{\noisy\imgB} & \pfig{\noisy\imgC} & \pfig{\noisy{\imgD}} & \pfig{\noisy{\imgE}}\\
        
         \hline \vspace{-8pt} \\ 
        \rotatebox{90}{\parbox[c]{2.3cm}{\centering CycleISP}} & \pfig{\cycle\imgA} & \pfig{\cycle\imgB} & \pfig{\cycle\imgC} & \pfig{\cycle{\imgD}} &
        \pfig{\cycle{\imgE}}\\
        
        \rotatebox{90}{\parbox[c]{2.3cm}{\centering DIP}} & \pfig{\dip\imgA} & \pfig{\dip\imgB} & \pfig{\dip\imgC} & \pfig{\dip{\imgD}} &
        \pfig{\dip{\imgE}} \\
        
        \rotatebox{90}{\parbox[c]{2.3cm}{\centering Noise2Void }} & \pfig{\ntov\imgA} & \pfig{\ntov\imgB} & \pfig{\ntov\imgC} & \pfig{\ntov{\imgD}} &
        \pfig{\ntov{\imgE}}\\
        
        \rotatebox{90}{\parbox[c]{2.3cm}{\centering Restormer }} & \pfig{\restormer\imgA} & \pfig{\restormer\imgB} & \pfig{\restormer\imgC} & \pfig{\restormer{\imgD}} &
        \pfig{\restormer{\imgE}} \\
        
        \rotatebox{90}{\parbox[c]{2.3cm}{\centering NAFnet }} & \pfig{\nafnet\imgA} & \pfig{\nafnet\imgB} & \pfig{\nafnet\imgC} & \pfig{\nafnet{\imgD}} &
        \pfig{\nafnet{\imgE}} \\

        \rotatebox{90}{\parbox[c]{2.3cm}{\centering Diffusion \\ \footnotesize (our)}} & \pfig{\concat\imgA} & \pfig{\concat\imgB} & \pfig{\concat\imgC} & \pfig{\concat{\imgD}} &
        \pfig{\concat{\imgE}} \\
        
        \rotatebox{90}{\parbox[c]{2.3cm}{\centering Diffusion +Text \\ \footnotesize (our)}} & \pfig{\cond\imgA} & \pfig{\cond\imgB} & \pfig{\cond\imgC} & \pfig{\cond{\imgD}} &
        \pfig{\cond{\imgE}} \\
        
        \rotatebox{90}{\parbox[c]{2.3cm}{\centering Clean S21 \\ \footnotesize (target)}} & \pfig{\gt\imgA} & \pfig{\gt\imgB} & \pfig{\gt\imgC} & \pfig{\gt{\imgD}} &
        \pfig{\gt{\imgE}} \\
        
    \end{tabular}
    \caption{Real-world denoising comparison of various methods applied to Samsung S21 camera captures. Our text-guided model achieves superior results compared to competing methods, including a non-text-guided diffusion model.}
    \label{fig:img_resultS21_more2}
\end{figure*}

\def\restormer#1{figs/allied/restormer/sample0#1.png}
\renewcommand{\noisy}[1]{figs/allied/noisy/sample0#1_low_res.png}
\renewcommand{\cycle}[1]{figs/allied/cycle/0000#1_raw4c.pt.png}
\renewcommand{\concat}[1]{figs/allied/loraconcat/sample0#1.png}
\renewcommand{\cond}[1]{figs/allied/loracond/sample0#1.png}
\renewcommand{\gt}[1]{figs/allied/gt/0000#1_raw4c.pt.png}

\renewcommand{\imgA}{00} 
\renewcommand{\imgB}{01} 
\renewcommand{\imgC}{02} 
\renewcommand{\imgD}{03} 
\renewcommand{\imgE}{04} 
\newcommand{\imgF}{17}

\renewcommand{\nafnet}[1]{figs/allied/NAFnet/sample0#1.png}
\newcommand{\tecdnet}[1]{figs/allied/TECDNet/sample0#1.png}

\begin{figure*}[p]
\vspace{-0.2in}
    \setlength\tabcolsep{2pt} %
    \renewcommand{\arraystretch}{0.86}
    \centering
    
    \begin{tabular}{ccccccc}
    
        \rotatebox{90}{\parbox[c]{2.3cm}{\centering Noisy Allied \\ \footnotesize (input)}} & \pfig{\noisy\imgA} & \pfig{\noisy\imgB} & \pfig{\noisy\imgC} & \pfig{\noisy{\imgD}} & \pfig{\noisy{\imgE}}& \pfig{\noisy{\imgF}}\\
        
         \hline \vspace{-8pt} \\ 
        \rotatebox{90}{\parbox[c]{2.3cm}{\centering CycleISP}} & \pfig{\cycle\imgA} & \pfig{\cycle\imgB} & \pfig{\cycle\imgC} & \pfig{\cycle{\imgD}} &
        \pfig{\cycle{\imgE}} &
        \pfig{\cycle{\imgF}}\\

        \rotatebox{90}{\parbox[c]{2.3cm}{\centering Restormer }} & \pfig{\restormer\imgA} & \pfig{\restormer\imgB} & \pfig{\restormer\imgC} & \pfig{\restormer{\imgD}} &
        \pfig{\restormer{\imgE}} &
        \pfig{\restormer{\imgF}} \\
        
        \rotatebox{90}{\parbox[c]{2.3cm}{\centering NAFnet }} & \pfig{\nafnet\imgA} & \pfig{\nafnet\imgB} & \pfig{\nafnet\imgC} & \pfig{\nafnet{\imgD}} &
        \pfig{\nafnet{\imgE}} &
        \pfig{\nafnet{\imgF}}\\

        \rotatebox{90}{\parbox[c]{2.3cm}{\centering TECDNet }} & \pfig{\tecdnet\imgA} & \pfig{\tecdnet\imgB} & \pfig{\tecdnet\imgC} & \pfig{\tecdnet{\imgD}} & \pfig{\tecdnet{\imgE}}& \pfig{\tecdnet{\imgF}} \\

        \rotatebox{90}{\parbox[c]{2.3cm}{\centering Diffusion \\ \footnotesize (our)}} & \pfig{\concat\imgA} & \pfig{\concat\imgB} & \pfig{\concat\imgC} & \pfig{\concat{\imgD}} &
        \pfig{\concat{\imgE}}  &
        \pfig{\concat{\imgF}} \\
        
        \rotatebox{90}{\parbox[c]{2.3cm}{\centering Diffusion +Text \\ \footnotesize (our)}} & \pfig{\cond\imgA} & \pfig{\cond\imgB} & \pfig{\cond\imgC} & \pfig{\cond{\imgD}} &
        \pfig{\cond{\imgE}}  &
        \pfig{\cond{\imgF}} \\
        
        \rotatebox{90}{\parbox[c]{2.3cm}{\centering Clean Allied \\ \footnotesize (target)}} & \pfig{\gt\imgA} & \pfig{\gt\imgB} & \pfig{\gt\imgC} & \pfig{\gt{\imgD}} &
        \pfig{\gt{\imgE}} &
        \pfig{\gt{\imgF}} \\
        
    \end{tabular}
    \caption{Real-world denoising comparison of various methods applied to Allied Vision Manta G-146C camera captures. Our text-guided model achieves superior results compared to competing methods, including a non-text-guided diffusion model.}
    \label{fig:img_resultSallied}
\end{figure*}

\end{document}